\documentclass[letterpaper, 10 pt, conference]{ieeeconf}  % Comment this line out if you need a4paper

\IEEEoverridecommandlockouts                              % This command is only needed if 
\usepackage{graphicx} % for pdf, bitmapped graphics files
\usepackage{array}
\usepackage{pifont}
\usepackage{xcolor} 
\usepackage{booktabs}  
\usepackage{multirow}
\usepackage{multicol}
\usepackage{url}
\usepackage{caption} 
\usepackage{cuted}
\usepackage{hyperref}

\hypersetup{
    colorlinks=false,        % Use boxes (borders) instead of colored text
    linkbordercolor={1 0 0}, % Red box for Figures, Tables, Sections
    citebordercolor={0 1 0}, % Green box for Citations [1]
    urlbordercolor={1 1 1}   % White (Invisible) box for URLs
}

\newcommand{\cmark}{{\color{green!70!black}\ding{51}}} % green checkmark
\newcommand{\xmark}{{\color{red}\ding{55}}}            % red cross

\title{\LARGE \bf
IndustryShapes: An RGB-D Benchmark dataset for 6D object pose estimation of industrial assembly components and tools
}

\author{Panagiotis Sapoutzoglou, Orestis Vaggelis, Athina Zacharia, Evangelos Sartinas, Maria Pateraki% <-this % stops a space 
\thanks{This work was funded by the HEU programme SOPRANO (GA No 101120990) and PANDORA (GA No 101135775). Special thanks to Stellantis—Centro Ricerche FIAT (CRF) for supporting the dataset collection.}
\thanks{All authors are with the National Technical University of Athens (NTUA), Greece (e-mail:\{psapoutzoglou, orestisvaggelis, azacharia, vsartinas, mpateraki\}@mail.ntua.gr)}}
\begin{document}
\bstctlcite{IEEEcontrol}

\maketitle

%%%%%%%%%%%%%%%%%%%%%%%%%%%%%%%%%%%%%%%%%%%%%%%%%%%%%%%%%%%%%%%%%%%%%%%%%%%%%%%%
% \begin{figure*}[!t]
%     \centering
%     \includegraphics[width=0.85\textwidth]{images/TEST.png}
%     \caption{Dataset sample images (bottom) overlaid with colored 3D models (top).}
%     \label{fig:dataset_samples}
% \end{figure*}

%\twocolumn[{%
%\renewcommand\twocolumn[1][]{#1}%

\begin{strip}
    \centering
    \vspace{-2cm}
    \captionsetup{type=figure}
    \includegraphics[width=\textwidth]{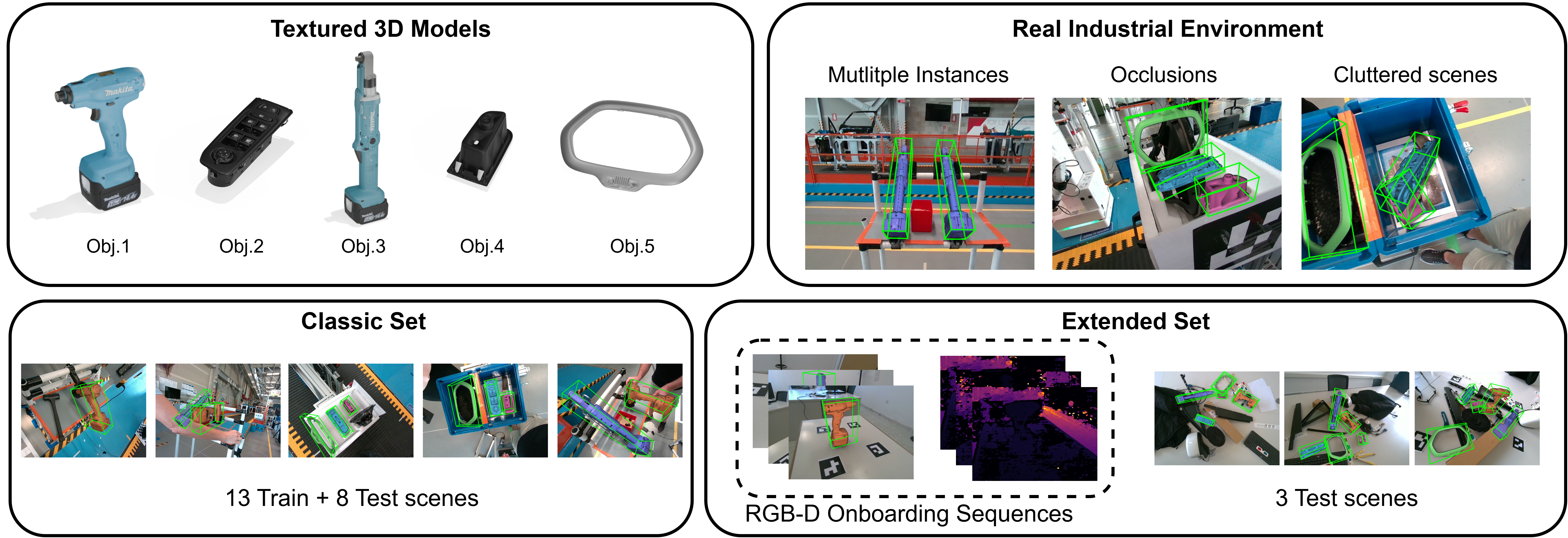}
    \captionof{figure}{Overview of the IndustryShapes dataset. Five industrial tools/components with challenging properties (weak/absent texture, symmetries, thin and reflective parts) are captured in realistic industrial environment scenes. Data are organized into two complementary sets: \emph{Classic} set for instance-level evaluation and  \emph{Extended} set tailored for novel-object methods, including RGB-D static onboarding sequences.}
    \label{fig:main}
\end{strip}%
%}]

\begin{abstract}

We introduce \emph{IndustryShapes}, a new RGB-D benchmark dataset of industrial tools and components, designed for both instance-level and novel object 6D pose estimation approaches. The dataset provides a realistic and application-relevant testbed for benchmarking these methods in the context of industrial robotics bridging the gap between lab-based research and deployment in real-world manufacturing scenarios. Unlike many previous datasets that focus on household or consumer products or use synthetic, clean tabletop datasets, or objects captured solely in controlled lab environments, \emph{IndustryShapes} introduces five new object types with challenging properties, also captured in realistic industrial assembly settings. The dataset has diverse complexity, from simple to more challenging scenes, with single and multiple objects, including scenes with multiple instances of the same object and it is organized in two parts: the \emph{classic set} and the \emph{extended set}. The \emph{classic set} includes a total of 4,6k images and 6k annotated poses. %3,7k images for training, along with \textasciitilde900 test images. %\emph{1361 synthetic images} and \emph{2746 real images} for training, along with \emph{1354 real test images}. 
The \emph{extended set} introduces additional data modalities to support the evaluation of model-free and sequence-based approaches. To the best of our knowledge, \emph{IndustryShapes} is the first dataset to offer RGB-D static onboarding sequences. We further evaluate the dataset on a representative set of state-of-the art methods for instance-based and novel object 6D pose estimation, including also object detection, segmentation, showing that there is room for improvement in this domain. 
The dataset page can be found in \href{https://pose-lab.github.io/IndustryShapes/}{\textcolor{cyan}{https://pose-lab.github.io/IndustryShapes}}.

\end{abstract}

%%%%%%%%%%%%%%%%%%%%%%%%%%%%%%%%%%%%%%%%%%%%%%%%%%%%%%%%%%%%%%%%%%%%%%%%%%%%%%%%

\section{INTRODUCTION}

6D pose estimation is a cornerstone of robotic perception, enabling robots to accurately estimate the position and orientation of objects relative to their camera(s). It is essential for robust manipulation in tasks such as pick and place and precise assembly/ disassembly and constitutes a foundational capability for perception-driven robotic systems in manufacturing industries. As modern factories evolve toward flexible automation, robust 6D pose estimation becomes increasingly important not only for repetitive tasks in robotic workcells but also for dynamic, collaborative environments. Despite its importance, 6D pose estimation in industrial environments remains challenging due to object and scene characteristics~\cite{Thalhammer_2024_TOR}.
Many tools and components are textureless, metallic, or reflective, while symmetric, thin and geometrically similar objects introduce ambiguities in pose estimation. Moreover, occlusions, clutter, variable and often unconstrained lighting conditions further degrade performance. Deep learning methods have advanced the field. However, the effectiveness of these methods depends on the availability of  diverse, and task-relevant data. Without realistic datasets, deep learning models often fail to generalize beyond clean, controlled environments. This limitation is especially pronounced in the industrial domain, where object properties and scene dynamics are particularly demanding. As emphasized in recent reviews~\cite{Thalhammer_2024_TOR} if the datasets’ complexity does not reflect the relevant real-world complexity of robotic scenarios, this will lead to a performance discrepancy at deployment. On the other hand an exhaustive annotation of all possible object configurations in complex scenes is impractical. While existing datasets like T-LESS~\cite{Hodan_2017_WACV}, ITODD~\cite{Drost_2017_ICCV}, or YCB-V~\cite{Xiang_2018_RSS} have contributed significantly to the field, they are often limited to controlled laboratory conditions or constrained domains such as bin-picking and household objects. These datasets are typically optimized for instance-level methods where object CAD models are known during training and broader support for novel object 6D pose estimation methods is underrepresented, including also modalities like onboarding sequences required for some of these methods.

% \begin{figure}
%     \centering
%     \includegraphics[width=0.5\textwidth]{images/samples/industryShapes_examples_mostly_crf.png}
%     % \caption{Dataset sample images (bottom) overlaid with colored 3D models (top).}
%     \label{fig:dataset_samples}
% \end{figure}
% \begin{figure}
%     \centering
%     \includegraphics[width=0.5\textwidth]{images/models_v3.png}
%     \caption{Overview of the dataset objects. 3D CAD model (top) and real-world image (bottom)%Each column shows the 3D CAD model (top) and a corresponding real-world image (bottom).
%     }
%     \label{fig:3d_models}
% \end{figure}

IndustryShapes addresses this critical gap by introducing an RGB-D dataset designed specifically for 6D pose estimation in industrial environments, forming a benchmark for evaluating developed methods under real-world conditions and bridging the gap between lab-based research and deployment in real-world manufacturing scenarios. It provides rich annotations and challenging object characteristics but also data formats that directly support modern pipelines, including model-based, model-free, and sequence-driven approaches.
The key contributions of the IndustryShapes dataset include:
\begin{itemize}
    \item Challenging object set in realistic industrial conditions: The dataset provides five industrial objects characterized by complex geometries, reflective surface, symmetries, and lack of texture, with the majority captured under realistic industrial conditions with varied complexity, from single-object setups to cluttered scenes involving multiple object instances and occlusions.
    \item Two complementary sets: the \emph{classic set}, designed primarily for instance-level methods, and the \emph{extended set}, tailored specifically for novel object pose estimation. Altogether, the dataset encompasses over 26k annotations. A unique feature of IndustryShapes is the inclusion of RGB-D static onboarding sequences, making it the first dataset explicitly developed to support model-free pose estimation methods.
    \item Benchmarking Framework: We evaluate IndustryShapes using representative state-of-the-art baselines for instance-level methods (e.g., EPOS~\cite{Hodan_2020_CVPR}, DOPE~\cite{Tremblay_2018_Corl},   ZebraPose~\cite{Su_2022_CVPR}) and latest novel object methods (e.g., FoundPose~\cite{ornek2024foundpose}, FoundationPose~\cite{foundationposewen2024}), as well as state-of-the-art object detection and segmentation methods (e.g., CNOS~\cite{nguyen2023cnos}, SAM-6D~\cite{Lin_2024_CVPR}). Our results demonstrate scope for advancement in current methodologies, reinforcing the dataset’s relevance as a robust benchmark for ongoing research.
\end{itemize}

\section{Related work}

\subsection{Datasets}

Diverse datasets have been developed and exploited in the past years to support research and benchmarking in 6D pose estimation across various application domains. The BOP-Classic-Core corpus, is a well-established collection that has served as the foundation for the BOP (Benchmark for 6D Object Pose Estimation) challenges since 2019 and comprises of seven datasets: LM-O~\cite{Brachmann_ECCV_2014}, T-LESS~\cite{Hodan_2017_WACV}, ITODD~\cite{Drost_2017_ICCV}, HB~\cite{Kaskman_2019_ICCV}, YCB-V~\cite{Xiang_2018_RSS}, IC-BIN~\cite{Doumanoglou_2016_CVPR}, and TUD-L~\cite{Hodan_2018_ECCV}. Complementing the core datasets, the BOP-Classic-Extra group includes datasets such as LM~\cite{Hinterstoisser_2013_ACCV}, HOPEv1~\cite{Lin_2021_IROS}, RU-APC~\cite{Rennie_2016_RAL}, IC-MI~\cite{tejani_2014_ECCV}, and TYO-L~\cite{Hodan_2018_ECCV}. These datasets extend the benchmark’s diversity by covering a broader range of object types and acquisition setups, primarily consumer and household objects and sequences designed to evaluate robustness to varying illumination, collected in both realistic and controlled laboratory environments.
The BOP-Industrial group expands the benchmark with datasets specifically tailored for industrial use such as bin picking and automated inspection. These include XYZ-IBD~\cite{xyzibd_dataset}, which features cluttered bin-picking scenes with occlusions and industrial parts with reflective and low-textured surfaces; ITODD-MV~\cite{Drost_2017_ICCV}, which extends ITODD scenes with additional images in a multiview setup; and IPD~\cite{Kalra_2024_CVPR}, which captures RGB, high-resolution depth and polarization images of industrial parts in realistic setups like trays and conveyors. These datasets capture key challenges encountered in industrial environments, such as weak textures, symmetric geometries, and lighting variability, though the focus is primarily on bin picking scenarios.

In addition to the above instance-specific datasets, recent efforts within the BOP benchmark have begun addressing the more challenging task of 6D pose estimation of unseen or novel objects, where systems must generalize beyond a predefined set of models known during training. Notable datasets supporting this paradigm include T-LESS, originally part of BOP-Classic-Core, which is often used in this context due to its inclusion of multiple similar-looking, texture-less industrial parts that test a model’s generalization capabilities. The HOPE dataset~\cite{Lin_2021_IROS} by NVIDIA includes 50 scenes of 28 toy grocery objects captured in household/office environments. The HANDAL dataset~\cite{Guo_2023_IROS} - as released in BOP- includes 40 objects from 7 categories of hardware tools and kitchen utensils. The HOT3D dataset~\cite{banerjee_2024_CVPR} stands out from others by focusing on egocentric, real-world hand-object interactions, capturing synchronized RGB and grayscale imagery from head-mounted devices. Outside the BOP benchmark, the Objectron dataset~\cite{Ahmadyan_2021_CVPR} provides annotated RGB-only video sequences of everyday objects captured from mobile devices in real-world settings, totaling 4M images, for category-level 3D object detection for AR applications. Moreover, the PACE dataset~\cite{Yang_2024_ECCV} includes 238 real-world objects, designed to advance the development and evaluation of pose estimation methods in cluttered scenarios.

Overall the majority of datasets target household objects and consumer goods (YCB-V, TUD-L, IC-MI, RU-APC, TYO-L, HOPE, HANDAL) supporting applications in domestic robotics and AR. Datasets like LM-O, IC-BIN, TUD-L, PACE and parts of HB adopt controlled laboratory setups, where lighting, background, and object arrangements are optimized for repeatable benchmarking under simplified or constrained conditions. Datasets that focus on industrial-relevant objects include T-LESS, ITODD, ITODD-MV, XYZ-IBD, IPD, and to a lesser extent HB, offering representative challenges such as weak texture, symmetries, and clutter. Notably, XYZ-IBD, IPD, and ITODD(-MV) are specifically designed to support bin-picking scenarios, featuring scenes with multiple instances and occlusions. Table~\ref{tab:datasets_overview} provides an overview of the datasets.

Unlike lab-captured datasets or those focused solely on bin-picking scenarios, IndustryShapes prioritizes industrial realism and challenging properties over object quantity, offering more general industrial assembly scenes featuring objects placed on holders, boxes, trolleys or in mixed-use environments that closely resemble real robotic workstations and human-robot collaboration settings. It introduces new object types with challenging properties, while also supporting research on novel object pose estimation by providing not only still images, but also RGB-D image sequences with close-range depth data and multiple objects per scene.

\begin{table*}[!ht]
    
    \centering
    \scriptsize
    \renewcommand{\arraystretch}{1.1}    
    \newcolumntype{C}[1]{>{\centering\arraybackslash}p{#1}}
     \caption{Overview of datasets for 6D object pose estimation}
     \label{tab:datasets_overview}
     %, highlighting key characteristics such as object categories, input modalities, presence of occlusions and textureless objects, scenes with multiple objects or instances, availability of onboarding videos, and targeted application domains.}
%     \begin{tabular}{|C{1.5cm}|C{1.7cm}|C{0.8cm}|C{0.6cm}|C{1.0cm}|C{0.8cm}|C{0.9cm}|C{0.9cm}|C{2.0cm}|}

\resizebox{0.99\textwidth}{!}{%
    \begin{tabular}{|C{1.5cm}|C{3.4cm}|C{2.0cm}|C{0.6cm}|C{1.0cm}|C{0.8cm}|C{0.9cm}|C{0.9cm}|C{3.5cm}|}
    
    \hline
        \textbf{Dataset} & \textbf{Object category} & \textbf{Input} & \textbf{Occl.} & \textbf{Textureless objects} & \textbf{Mult. objects} & \textbf{Mult. instance} & \textbf{Onboard. videos} & \textbf{Application domain} \\ \hline
        LM-O~\cite{Brachmann_ECCV_2014} & 8 household, workshop tools & RGB-D & \cmark & \cmark & \cmark & \xmark & \xmark & general rob., AR \\ \hline
        T-LESS~\cite{Hodan_2017_WACV} & 30 industry-relevant  & RGB-D & \cmark & \cmark & \cmark & \cmark & \xmark & industrial rob. \\ \hline
        ITODD-MV~\cite{Drost_2017_ICCV} & 28 industry-relevant & Gray-D & \cmark & \cmark & \cmark & \cmark & \xmark & industrial rob. (bin picking) \\ \hline
        HB~\cite{Kaskman_2019_ICCV} & 33 objects & RGB-D & \cmark & \cmark & \cmark & \xmark & \xmark & AR, general rob. \\ \hline
        YCB-V~\cite{Xiang_2018_RSS} & 21 household objects & RGB-D & \cmark & \xmark & \cmark & \cmark & \xmark & general rob., AR \\ \hline
        ICBIN~\cite{Doumanoglou_2016_CVPR} & 2 objects & RGB-D & \cmark & \cmark & \cmark & \cmark & \xmark & robotics (bin picking) \\ \hline
        TUD-L~\cite{Hodan_2018_ECCV} & 3 household objects & RGB-D & \xmark & \cmark & \xmark & \xmark & \xmark & general rob. - light \\ \hline
        IC-MI~\cite{tejani_2014_ECCV} & 6 household objects & RGB-D & \cmark & \cmark & \cmark & \cmark & \xmark & general rob., AR \\ \hline
        RU-APC~\cite{Rennie_2016_RAL} & 14 consumer products & RGB-D & \cmark & \xmark & \cmark & \cmark & \xmark & robotics in warehouse  \\ \hline
        TYO-L~\cite{Hodan_2018_ECCV} & 21 household objects & RGB-D & \xmark & \xmark & \xmark & \xmark & \xmark & general rob. - light \\ \hline
        IPD~\cite{Kalra_2024_CVPR} & 22 industrial parts & RGB-D \& polar. & \cmark & \cmark & \cmark & \cmark & \xmark & industrial rob. (bin picking) - light \\ \hline
        XYZ-IBD~\cite{xyzibd_dataset} & 17 industrial parts & Gray-D & \cmark & \cmark & \cmark & \cmark & \xmark & industrial rob. (bin picking) - light \\ \hline
        HOPE~\cite{Lin_2021_IROS} & 28 toy grocery objects & RGB-D & \cmark & \xmark & \cmark & \cmark & \cmark & general rob., AR \\ \hline
        HANDAL~\cite{Guo_2023_IROS} & 40 hardware tools, utensils & RGB-D & \cmark & \cmark & \cmark & \xmark & \cmark & general rob. (part. industrial) \\ \hline
        HOT3D~\cite{banerjee_2024_CVPR} & 33 household objects & RGB & \cmark & \xmark & \cmark & \cmark & \cmark & AR, egocentric hand-object inter. \\ \hline
        Objectron~\cite{Ahmadyan_2021_CVPR} & 9 categories (mult) & RGB & \xmark & \xmark & \xmark & \xmark & \xmark & AR, 3D object detection \\ \hline
        PACE~\cite{Yang_2024_ECCV} & 238 objects & RGB-D & \cmark & \cmark & \cmark & \xmark & \xmark & general rob., AR \\ \hline
        IndustryShapes & 5 industrial objects & RGB-D & \cmark & \cmark & \cmark & \cmark & \cmark & industrial rob. \\ \hline
    \end{tabular}
   }
\end{table*}

\subsection{Methods}
\label{sec:sota_methods}

6D object pose estimation has witnessed rapid progress in the last years, especially with the advent of deep learning. In contrast to the surveys of~\cite{Fan_2022_ACMCS, Marullo_2023_MTA} that primarily focused on instance-level methods, Liu et al.~\cite{Liu_2024_ARXIV} systematically covers instance-level, category-level, and unseen object pose estimation and provides a multidimensional analysis of methods considering input modalities (RGB, RGBD, depth), object properties (textureless, symmetric, transparent, occluded), inference mode stages (e.g. segmentation, correspondence prediction, template matching, pose regression, pose refinement). It also reviews evaluation criteria according to the pose DoF (3DoF, 6DoF, 9DoF).

\textbf{Instance-level 6D object pose estimation.} Here, we provide a brief overview of relevant recent deep learning-based methods, categorizing them into \emph{correspondence-based} and \emph{direct pose regression} methods, highlighting representative examples in each. For a broader and more detailed analysis of the full spectrum of 6D pose estimation research, the reader is referred to recent dedicated surveys. \newline
\emph{Correspondence-based methods} predict 2D–3D associations between the image pixels and points on the object’s 3D model, typically followed by a Perspective-n-Point (PnP) solver to recover the pose. Keypoint-based methods, a specific case of these approaches, focus on predicting the 2D image projections of a small set of predefined 3D keypoints, as done in DOPE~\cite{Tremblay_2018_Corl}. PVNet~\cite{Peng_2019_CVPR} introduced pixel-wise voting mechanisms for keypoint localization aiming to increase the number and quality of point correspondences under partial occlusions. 
Dense correspondence-based approaches generalize this idea by predicting either visible 3D object coordinates for each pixel, as in Pix2Pose~\cite{Park_2019_ICCV}, or multiple correspondence hypotheses per pixel, as in EPOS~\cite{Hodan_2020_CVPR}.%, to better manage symmetries and occlusions. 
~ZebraPose \cite{Su_2022_CVPR} further introduces a coarse-to-fine hierarchical surface encoding.\newline
\emph{Direct pose regression methods} attempt to predict the 6D pose parameters directly from an input image. Early works such as PoseCNN~\cite{Xiang_2018_RSS} regressed translation vectors and discretized rotation angles into classification bins, while SSD-6D~\cite{Kehl_2019_ICCV} integrated viewpoint classification into a detection framework. Methods like GDR-Net~\cite{Wang_2021_CVPR} incorporate geometric guidance into the regression process, using dense feature representations to improve robustness and accuracy. Despite their simplicity, direct regression methods historically struggled with rotation representation, instabilities and ambiguities, caused by symmetries and occlusions, often leading to reduced accuracy compared to correspondence-based approaches~\cite{Liu_2024_ARXIV}. 

\textbf{Novel object 6D pose estimation} has emerged only recently as a distinct and challenging subfield and enables pose prediction for entirely new objects without retraining and can be categorized as either model-based or model-free. \emph{Model-based methods}~\cite{ornek2024foundpose, Lin_2024_CVPR, foundationposewen2024, labbe2022megapose, nguyen2024gigaPose, osop2022, caraffa2024freeze, Ausserlechner_2024_ICRA} typically use a 3D CAD model of the object at test time, rendering it from multiple viewpoints to generate templates that represent different poses. These templates are then compared to the observed scene to identify the best alignment with the object’s viewpoint, using feature descriptors, image similarity, or 3D point clouds in the case of RGB-D or depth input. In contrast, \emph{model-free methods}~\cite{sun2022onepose, he2022oneposeplusplus, liu2022gen6d, bundlesdfwen2023} do not rely on explicit 3D representations but instead operate on reference images or videos of the object. They often employ feature matching, visual localization techniques to build an internal object representation. Model-free pipelines typically run in two phases: onboarding and inference. During onboarding, the system observes a short sequence of the novel object, the so-called onboarding sequence, and constructs a representation using methods such as SfM, or neural rendering. In the inference phase, this learned representation is used to estimate the object’s 6D pose from a single frame. %\TODO{In the most challenging and purest form of model-free pose estimation, the system must handle entirely unseen objects with no prior exposure to the 2D instance or category and without relying on pretrained 2D detectors, thereby excluding conventional object detection from the pipeline.}

\textbf{Object Detection and Segmentation.} Recent advancements in foundation models have significantly influenced novel object segmentation. Many pipelines for pose estimation now rely on a preliminary detection or segmentation stage. Methods like CNOS~\cite{nguyen2023cnos} leverage the Segment Anything Model (SAM)~\cite{kirillov2023segany} to generate numerous segmentation proposals from an RGB image. These proposals are then matched against pre-rendered templates of a given CAD model by comparing feature descriptors extracted by DINOv2~\cite{oquab2023dinov2}. This approach allows for training-free segmentation of novel objects specified by their CAD models. Similarly, SAM-6D  adapts this approach for RGB-D data, where its Instance Segmentation Model (ISM) uses SAM to generate initial object proposals. However, it introduces a more detailed object matching score that evaluates proposals based on a combination of semantics, appearance, and geometric properties to identify valid instances of novel objects in cluttered scenes.

\section{The IndustryShapes dataset}

\subsection{Objects and 3D models}
%The dataset provides high-quality 3D CAD models (see Fig.\ref{fig:3d_models}) of industrial tools (\emph{Objects} 1 and \emph{3}) and components (\emph{Objects} 2, 4, and 5) found in assembly line workstations. All objects models are scaled to millimeters. The objects exhibit diverse challenging characteristics: lack of texture, complex geometries, reflective surfaces, symmetries and thin structures.

The dataset provides high-quality 3D CAD models (see Fig.~\ref{fig:main}) of industrial tools (Objects 1 and 3) and components (Objects 2, 4, and 5) found in assembly line workstations.
Furthermore, these models include photorealistic textures allowing for the generation of synthetic data in complex scenes with distractor objects using rendering tools like BlenderProc~\cite{Denninger2023}. The objects were deliberately selected to exhibit a range of challenging characteristics that are underrepresented in existing benchmarks: lack of texture, complex geometries, reflective surfaces, thin structures, and symmetries. This focus on challenging, domain-relevant properties provides a more demanding test for modern algorithms than many larger-scale datasets featuring simpler household objects, a fact reflected in our benchmark performance analysis in \ref{sec:performance_eval}.

% \begin{figure*}[htbp]
%     \centering
%     \includegraphics[width=0.8\textwidth]{images/models_v3.png}
%     \caption{Overview of the dataset objects. 3D CAD model (top) and real-world image (bottom)%Each column shows the 3D CAD model (top) and a corresponding real-world image (bottom).
%     }
%     \label{fig:3d_models}
% \end{figure*}

%\panos{The dataset provides high-quality 3D CAD models (see Fig.\ref{fig:3d_models}) of industrial tools (Objects 1 and 3) and components (Objects 2, 4, and 5) found in assembly line workstations. All objects are scaled to millimeters. The objects present various characteristics. The majority of them lack significant texture (Objects 2,3, and 4), while others exhibit reflective surfaces (Object 5), making RGB-based localization methods less effective. Objects 4 and 5 feature complex geometries, including hollow and skeletal structures. These properties pose significant challenges for perception and robotic manipulation.}

\subsection{Dataset structure and Composition} 
\label{dataset_structure}
The dataset comprises of two sets: the \emph{classic set} and the~\emph{extended set}. The~\emph{classic set} was originally created to support instance-level pose estimation methods and includes a variety of real and lab captured scenes. 
The real scenes were primarily captured in a realistic industrial assembly setting under realistic lighting conditions and feature varying scene complexities, from single-object setups to cluttered scenes with multiple objects, occurring instances of the same object, and significant occlusions. These were complemented with images of single objects acquired in laboratory conditions using a turn-table setup and systematically sampling views of the object using a fixed camera configuration~\cite{Papadaki_2023_jimaging}. Data in both industrial and laboratory settings were captured using the Intel RealSense D455 RGB-D camera at a resolution of 640 x 480, selected to balance spatial detail with real-time processing demands in robotic environments, while satisfying the minimum depth distance of the sensor, i.e. ~0.52 m.
% and further downsampled to 640 x 480 pixels.
In addition, synthetic images were generated using an OpenGL-based rendering pipeline \cite{Pateraki_2023_IJCCVICGTA}, exploiting the photorealistic object texture to produce RGB-D data.

\begin{figure*}[htbp]
    \centering
    \includegraphics[width=\textwidth]{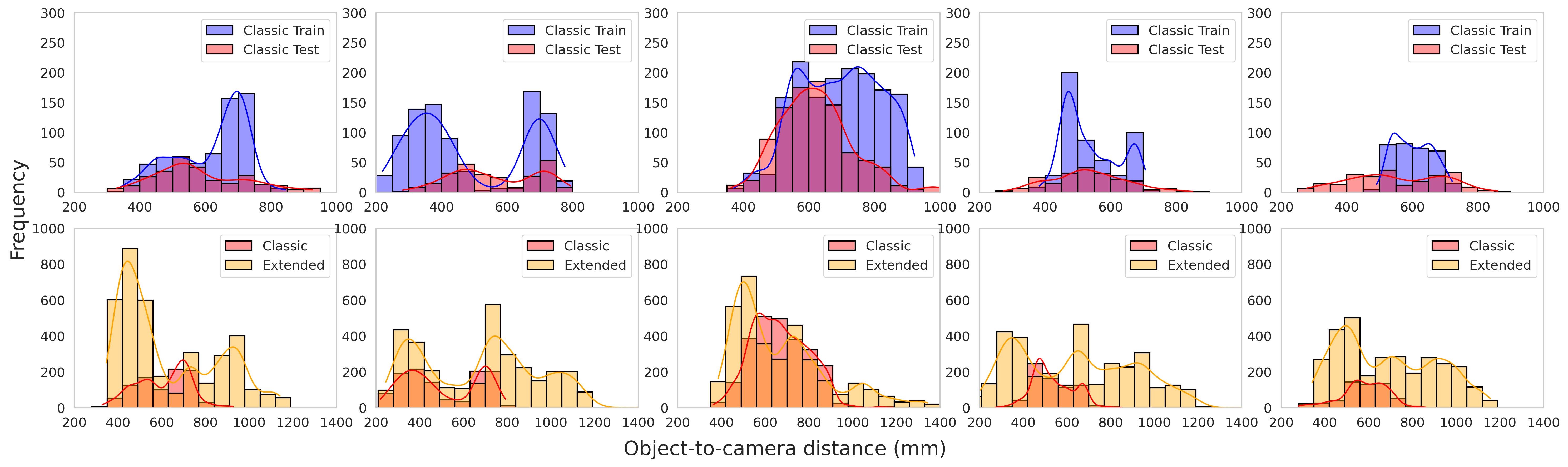}
    \caption{Distribution of object-to-camera distances for annotated poses, grouped by object (1 to 5 from left to right). Top row: annotated poses in the training (blue) and  test (magenta) data of the \emph{classic set}. Bottom row: annotated poses of the \emph{classic set} (orange) and the \emph{extended set} (yellow).}
    \label{fig:classic_vs_extended_distance_hists}
\end{figure*}

\textbf{Classic set}.~The \emph{classic set} includes a total of 21 scenes, with 13 scenes used for training and 8 scenes for test. The training data include 1217 images from 4 scenes captured in laboratory conditions and 1122 images from 8 scenes captured from the industrial assembly setting. All scenes with the exception of one from this setting feature a single object per scene. In addition to these, 1361 synthetically rendered images for Object 3, are included in the training data, totaling 3700 frames. The test set consists of 923 images from challenging scenes in the real industrial environment, featuring occlusions, multiple object instances and diverse spatial configurations. In these scenes, objects may appear in different physical locations or in different configurations with other objects, in boxes, trays or on tool holders. The rationale for using primarily single-object scenes for training is that it enables accurate supervision for learning accurate object-specific 6D pose representations without interference from background clutter, while testing on challenging world scenes enables robust evaluation of a model's generalization ability under realistic deployment conditions. Furthermore,  this design lifts practical constraints as capturing and annotating all possible object configurations and placements in real industrial environments is time-consuming and impractical. In total, the \emph{classic} set contains 4623 images and approximately 6k annotated poses (see Table~\ref{tab:dataset-groups}). The annotations per object and the number of train and test scenes depicting each object are listed in Table~\ref{tab:annotations-per-object}. Scenes feature object-to-camera distances from 200 mm to 1000 mm, with the majority falling within the 400 to 800 mm range (Fig.~\ref{fig:classic_vs_extended_distance_hists}). This range reflects typical distances used in assembly lines to enable robotic manipulation, while adhering to safety regulations that minimize the risk of collisions with the surrounding environment. Furthermore, specific objects had characteristic placements and orientations within the workstation setups, or were observed from specific viewpoints, as the case of objects placed in boxes, which introduced visibility constraints.  %\TODO{For example, Object 2 and 4 was rarely observed from viewpoints near the equator, as it was typically placed in boxes which introduced visibility constraints.}

\begin{table}[htbp]
    \caption{Dataset groups}
    \label{tab:dataset-groups}
    \centering
    \begin{tabular}{@{}lcc@{}}
        \toprule
        \textbf{Group} & \textbf{Annot.} & \textbf{Scenes} \\
        \midrule
        Classic Train     & 3,9k  & 13\\
        Classic Test      & 1,9k  & 8 \\
        Onboarding        & 6,3k  & 10 \\
        Extended Test     & 10.3k & 3 \\
        \midrule
        \textbf{Total}    &  22.4k & 34 \\
        \bottomrule
    \end{tabular}
\end{table}
\vspace{-0.3 cm}
\begin{table}[htbp]
    \caption{Annotations per object \& number of scenes per object. TR: Train, TE: Test, ON: Onboarding}
    \label{tab:annotations-per-object}
    \centering
    \begin{tabular}{@{}lccccc@{}}
        \toprule
        \textbf{Object} & \textbf{Annot.} & \multicolumn{2}{c}{\textbf{Classic Sc.}} & \multicolumn{2}{c}{\textbf{Extended Sc.}}\\
        & & TR & TE & ON & TE\\
        \midrule
        Obj. 1 & 4,6k & 3 & 2 & 2 & 3\\
        Obj. 2 & 4.3k & 5 & 5 & 2 & 3\\
        Obj. 3 & 6k   & 2 & 2 & 2 & 3\\
        Obj. 4 & 3.9k & 2 & 4 & 2 & 3\\
        Obj. 5 & 3.6k & 2 & 5 & 2 & 3\\
        \midrule
        \textbf{Total} & 22.4k & 13 & 8 & 10 & 3\\
        \bottomrule
    \end{tabular}
\end{table}

\begin{figure*}[htbp]
    \centering
    \includegraphics[width=\textwidth]{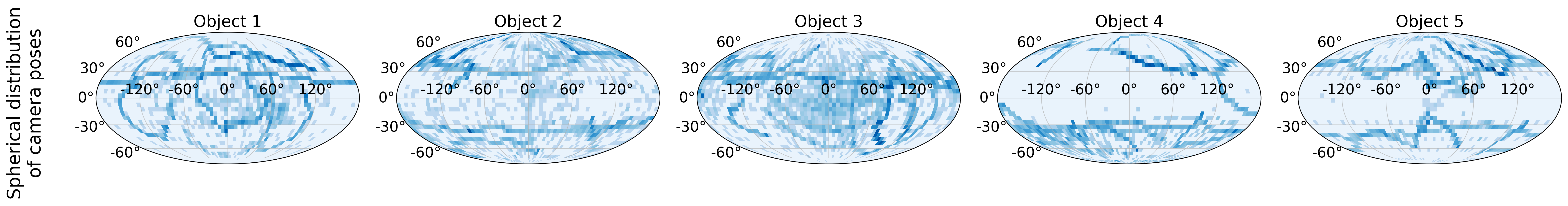}
    \caption{Pose distribution per Object. Visualization of the overall spherical viewpoint coverage of the complete IndustryShapes dataset in Mollweide projection, indicating the density and pose variation. }
    \label{fig:pose_distributions}
\end{figure*}

\textbf{Extended set}.The \emph{extended set} was introduced to support the benchmarking for novel object pose estimation methods, both model-based and model-free methods, as they gained momentum. The set includes RGB-D static onboarding sequences, two per object, with different placements of the objects on a table. One where the object is resting naturally upright on the table and another where the object is laid flat or positioned on its opposite side or rotated to expose different parts of the object. The set also includes three test scenes featuring all five objects in an office environment with unconstrained lighting, various distractors, occlusions, acquired from diverse viewpoints. Combined with the objects’ challenging characteristics, these factors make the pose estimation task particularly demanding. The sequences of the set were captured, at a resolution of 640×480 pixels, using a hand-held Intel RealSense D405 RGB-D camera, offering close-range depth information, necessary for the onboarding sequences. It is important to note that, to the best of our knowledge, IndustryShapes is the first dataset that provides RGB-D static onboarding sequences. The 10 onboarding sequences (2 per object) of the set, provide in average 650 frames each and a total of ~6.3k RGB-D frames, while the 3 test scenes include over 2k images and approximately 10.3k annotated object instances. Together with the onboarding sequences, they significantly enrich the dataset, providing dense camera orientations (Figure~\ref{fig:pose_distributions}) and well distributed object-to-camera distances (Figure~\ref{fig:classic_vs_extended_distance_hists}). 

All data include RGB images, depth maps, annotated 6D object poses, segmentation masks, and are formatted according to the BOP specifications. The annotation procedure is detailed in Sec.~\ref{sec:Annotation}.

\iffalse
\subsection{Synthetic image generation}
\label{sec:synthetic}
To enrich the training dataset, synthetic images can be generated using a custom OpenGL-based rendering pipeline. Objects are systematically rendered against a uniform background by placing each 3D model at the center of a virtual hemisphere and sampling viewpoints in spherical coordinates $\phi$ and $\theta$, with the azimuthal angle $\phi \in [0, 360)$ degrees and the polar angle $\theta \in [5, 90]$ degrees. Radial distances are sampled in the range of $600$ to $800$ mm. %Views were densely sampled around the hemisphere in radial %distances ranging from $600 - |\min t_z|$ to $800 + |\min t_z|$ where $t_z$ is 
%distances ranging from $600$ to $800$. The azimuthal angle $\phi\in[0,360)$ degrees, whereas the polar angle $\theta\in[5,90]$.% was in the range of 5-90 degrees. 
To ensure dense coverage of the hemisphere, viewpoints are sampled at angular intervals of $5^\circ$ in $\phi$ and $10^\circ$ in $\theta$.
%~The step interval for consecutive views was set to 5 and 10 degrees for $\phi$ and $\theta$, respectively. 
To enhance diversity, additional random translation and rotation can be applied to each pose. These included translations in mm within the ranges $t_x\in(-200,200), t_y\in(-100,100) $ and $t_z\in(-100,100)$, and rotations in degrees $r_x\in(-20.5,10.5),r_y\in(-1.5,1.5)$ and $r_z\in(-1.5,1.5)$. The synthetic image and pose generation framework is open-source ~\cite{6dPoseGenerator_2025repo} and comes with parameters customization (see suppl.).
\fi

\subsection{6D pose annotation of real image sequences}
\label{sec:Annotation}

Given the diverse types of data in the dataset %(see sec. \ref{sec:IndustryShapes_classic})
, different approaches were employed regarding the annotation process. For the lab-captured data, the turn-table setup enabled the use of a marker-based approach, where camera poses were estimated from ArUco markers (see \cite{Papadaki_2023_jimaging}). Similarly, in the industrial environment, the marker-based method was applied wherever marker installation was feasible. However, in cases where marker placement was not possible, a semi-automatic annotation pipeline was developed using the Structure-from-Motion (SfM) software~\cite{agisoft2025metashape}. In this setup, SfM reconstruction was combined with manually defined anchor points corresponding to known 3D coordinates from the CAD model of the object. These anchor points enabled the establishment of 2D–3D correspondences between the object in the images and its 3D model. The software automatically tracked the 2D marker locations across frames using the computed camera poses. These 2D points, together with the corresponding 3D coordinates, were then used to solve the PnP problem, yielding the object’s pose relative to the camera. While this method requires less manual effort and supervision, its accuracy is comparable to fully manual pose annotation. % \textcolor{red}{it consistently produces highly accurate pose annotations}. 
Considerable effort was devoted in ensuring annotation quality, as inaccurate ground truth poses can result in erroneous evaluations. The accuracy of the annotations on real images was evaluated by comparing captured depth data ($d_c$) with rendered depth images ($d_r$) from the ground truth poses. After filtering for valid pixels and excluding outliers, the absolute depth error between the annotated poses and the captured depth was found to be less than 12 mm for the classic set and approximately 5 mm for the extended set. This represents a relative error of less than 5\% when compared to the objects mean diameter of 254 mm.

\begin{table}[ht]
\caption{Depth difference statistics of the $\delta = d_c - d_r$ distribution (in mm)}
\label{tab:depth_stats}
\centering
% \resizebox{0.9\linewidth}{!}{%
\begin{tabular}{@{}l|llllc@{}}
    \toprule
    \textbf{} & $\mu_{\delta}$ & $\sigma_{\delta}$ &
    $\mu_{|\delta|}$  &
    $med_{|\delta|}$ \\
    \midrule
    \emph{Classic} (D455) & -1.85 & 11.6 & 9.3 & 7.15  \\
    \emph{Extended} (D405) & -1.01 & 5.2 & 10.2 & 8.95  \\
     
    \bottomrule
\end{tabular}
% }
\end{table}

% \begin{figure*}[htbp]
%     \centering
%     \includegraphics[width=0.8\textwidth]{images/Test_scenes.png}
%     \caption{Test scenes in the classic set. }
%     \label{fig:classic_test}
% \end{figure*}

\section{Benchmark}
\label{sec:benchmark}
\subsection{Evaluation metrics}
%\vaggelis{The accuracy and precision of the 6D object pose estimation are evaluated through the comparison between the predicted and the known ground truth pose of the depicted object. There is a well-established strategy in evaluating the 6D object pose estimation in the literature \cite{hodan2020bop} that involves several metrics. The most widely used metrics are: (i) Average Distance for distinguishable (ADD or ADD(-S)),% symmetric distinguishable (ADD-S), and indistinguishable (ADI) objects, 
%which calculates the average distance between the vertices in the 3D space. (ii) Visible Surface Discrepancy (VSD), which defines the object location using distance maps between the input and rendered image, while being invariant to symmetries and encountering only the visible object parts. (iii) Maximum Symmetry-Aware Surface Distance (MSSD) which computes the maximum distance and is independent of the object geometry. (iv) Maximum Symmetry-Aware Projection Distance (MSPD) that is similar to MSSD but for the 2D space, leaving out the Z-axis misalignment. \athena{v) Average Recall (AR) which is calculated as the mean of MSPD, MSSD and VSD. }These are the evaluation metrics we utilized in our work.}
We adhere to the BOP challenge protocol~\cite{hodan2020bop}, evaluating methods using four pose error metrics based on the estimated pose and the ground-truth pose: the Visible Surface Discrepancy (VSD), computing differences from renderings of the estimated and ground-truth poses only over the visible surface areas; the Maximum Symmetry-Aware Surface Distance (MSSD), measuring the maximum surface distance considering object symmetry; the Maximum Symmetry-Aware Projection Distance (MSPD) measuring the maximum 2D projection error in pixels, considering symmetry; and the Average Distance for distinguishable (ADD) objects which quantifies the average misalignment between the model’s vertices in the true and estimated pose. Although not symmetry-aware, ADD is widely used in prior literature and provides a valuable reference point for evaluating pose accuracy. In line with the BOP protocol, the Average Recall (AR), used to summarize the overall performance, is computed as the mean recall of VSD, MSSD, and MSPD. In addition to pose estimation metrics, we also report the mean Average Precision (mAP) for both detection and segmentation, following common practice in the BOP challenge.
% \vspace{-0.3cm}
\subsection{Baselines and evaluation setup}
\label{sec:eval_setup}

We evaluate object detection, segmentation and pose estimation methods using both instance-level and novel object baselines. For instance-level methods, we deliberately selected EPOS~\cite{Hodan_2020_CVPR}, ZebraPose~\cite{Su_2022_CVPR}, and DOPE~\cite{Tremblay_2018_Corl}. EPOS was chosen for its efficiency and robustness, as it handles object symmetries better than early BOP baselines such as Pix2Pose~\cite{Park_2019_ICCV}, supports multi-object training within a unified network %unlike per-object approaches, 
and remains competitive in the BOP benchmark, particularly under the VSD and MSSD metrics. ZebraPose was included as a state-of-the-art representative from the 2023 BOP challenge and DOPE was selected as a representative keypoint-based approach. In contrast, we excluded detector-dependent methods such as GDRNet~\cite{Wang_2021_CVPR}, whose accuracy is tightly coupled to bounding-box quality and whose per-object training time made it impractical for our benchmarking scope, as reported in~\cite{Thalhammer_2024_TOR}. All selected instance-level methods were trained and tested on the Classic IndustryShapes dataset. For novel-object methods, we included FoundPose~\cite{ornek2024foundpose} (a model-based approach) and FoundationPose~\cite{foundationposewen2024} (supporting both model-based and model-free setups). We chose to exclude other methods such as Gen6D~\cite{liu2022gen6d} due to the significant manual effort required to adapt it to custom objects, and OnePose++~\cite{he2022oneposeplusplus} as it relies on an iOS-only mobile app for data capture that does not scale well to large datasets. Both selected methods were evaluated directly on the Classic and Extended datasets without retraining. Publicly available code was used for all methods.%, with modifications tailored to the experimental use case.

For object detection, we include CNOS~\cite{nguyen2023cnos}, which leverages contrastive learning to generalize to unseen objects without retraining. For segmentation, we adopt SAM-6D~\cite{Lin_2024_CVPR}, a recent approach that combines the Segment Anything Model (SAM) with a 6D pose estimation head, enabling robust segmentation-driven object localization.

\begin{table*}[ht]
\caption{Performance of baseline methods on IndustryShapes dataset, \emph{classic} and \emph{extended} sets. The table shows Average Recall (AR) scores per method based on \cite{hodan2020bop}, along with the recall rates for the ADD, MSPD, MSSD and VSD error metrics.  %Comparison of pose estimation metrics across different methods on IndustryShapes Classic and Extended datasets.
}
\label{tab:pose_metrics}
\centering
\resizebox{0.85\textwidth}{!}{%
\begin{tabular}{llccccc|ccccc}
\toprule
\multicolumn{2}{c}{} & \multicolumn{5}{c|}{\textbf{IndustryShapes Classic}} & \multicolumn{5}{c}{\textbf{IndustryShapes Extended}} \\
\cmidrule(lr){3-12} %\cmidrule(lr){8-12}
\multicolumn{2}{c}{} &  \textbf{ADD} & \textbf{MSPD} & \textbf{MSSD} & \textbf{VSD} & \textbf{AR}   & \textbf{ADD} & \textbf{MSPD} & \textbf{MSSD} & \textbf{VSD} & \textbf{AR} \\
\midrule
\multirow{3}{*}{\rotatebox[origin=c]{90}{\shortstack{\textbf{Inst.}\\\textbf{level}}}}  
  & EPOS       &   0.64    &    0.54   &    0.56   &   0.41    &   \textbf{0.51}    &   -    &    -   &   -    &   -    &    -   \\
  & DOPE       &  0.13   &  0.09   &  0.09     &   0.05    &   0.08    &   -    &    -   &   -    &   -    &    -   \\
  & Zebrapose  & 0.61  & 0.54  & 0.55  & 0.40  & 0.50  &   -    &   -    &    -   &   -    &   -    \\ [.1cm]
\midrule
\multirow{3}{*}{\rotatebox[origin=c]{90}{\shortstack{\textbf{Novel}\\\textbf{objects}}}}
  & FoundPose   &   0.55    &   0.31    &   0.39    &    0.22   &    0.30   &   0.34   &   0.41    &    0.25   &    0.18  &   0.28  \\
  & FoundationPose (MB)      &   0.78    &    0.73   &    0.74   &    0.53  &    \textbf{0.67}    &    0.81   &   0.83    &    0.74   &    0.50   &    \textbf{0.69}\\
  & FoundationPose (MF)       &   0.44    &    0.26   &    0.31   &    0.18   &   0.25    &    0.48   &   0.40    &    0.37   &    0.22   &   0.33 \\
\bottomrule
\end{tabular}
}
\end{table*}

\begin{table*}[ht]
\caption{Average Recall (AR) per object per baseline method across the IndustryShapes dataset.}
\label{tab:ar_per_object}
\centering
\resizebox{0.85\textwidth}{!}{%
\begin{tabular}{llccccc|ccccc}
\toprule
\multicolumn{2}{c}{} & \multicolumn{5}{c|}{\textbf{IndustryShapes Classic}} & \multicolumn{5}{c}{\textbf{IndustryShapes Extended}} \\
\cmidrule(lr){3-12}
\multicolumn{2}{c}{} & \textbf{Obj 1} & \textbf{Obj 2} & \textbf{Obj 3} & \textbf{Obj 4} & \textbf{Obj 5} 
                    & \textbf{Obj 1} & \textbf{Obj 2} & \textbf{Obj 3} & \textbf{Obj 4} & \textbf{Obj 5} \\
\midrule
\multirow{3}{*}{\rotatebox[origin=c]{90}{\shortstack{\textbf{Inst.}\\\textbf{level}}}}  
  & EPOS       & 0.66 & \textbf{0.26} & 0.59 & \textbf{0.25} & 0.00 & - & - & - & - & - \\
  & DOPE       & 0.18 & 0.05 & 0.03 & 0.10 & 0.00 & - & - & - & - & - \\
  & Zebrapose  & \textbf{0.84} & 0.11 & 0.49 & \textbf{0.39} & \textbf{0.11} & - & - & - & - & -  \\ [.1cm]
\midrule
\multirow{3}{*}{\rotatebox[origin=c]{90}{\shortstack{\textbf{Novel}\\\textbf{objects}}}}  
  & FoundPose           & 0.43 & 0.34 & 0.25 & 0.11 & 0.15 & 0.48 & 0.17 & 0.13 & 0.23 & 0.04 \\
  & FoundationPose (MB) & \textbf{0.78} & \textbf{0.49} & \textbf{0.64} & \textbf{0.47} & \textbf{0.31} & \textbf{0.79} & \textbf{0.54} & \textbf{0.42} & \textbf{0.54} & \textbf{0.56} \\
  & FoundationPose (MF) & 0.53 & 0.32 & 0.12 & 0.09 & 0.23 & 0.47 & 0.26 & 0.07 & 0.02 & 0.38 \\[.1cm]
\bottomrule
\end{tabular}
}
\end{table*}

%\vspace{-0.2cm}
\subsection{Performance evaluation}
\label{sec:performance_eval}
%\vspace{-0.1cm}
\paragraph{6D Pose Estimation}
The results, summarized in Table~\ref{tab:pose_metrics}, provide a direct comparison of the baseline methods across the diverse object types and varying scene complexities. %under both instance-based and novel object frameworks. 
%The evaluation highlights each method’s performance across the diverse object types and varying scene complexities. %present in the IndustryShapes dataset, offering insight into their relative strengths and weaknesses in real-world 6D pose estimation tasks. 
The challenging nature of the dataset is evident from the generally low performance across most methods. Notably, the strength of recent novel object pose estimation approaches is emphasized, as they achieve comparable performance to instance-level methods. %—despite not requiring retraining on the target dataset.
In addition, an analysis of the AR per object reveals significant variability in performance depending on the specific object and scene context (Table~\ref{tab:ar_per_object}). This is evident in the novel-object case, where no training is performed on the target objects. Certain objects consistently prove to be more challenging than others. For instance, Object 5, which is reflective with a thin structure, shows significantly lower performance compared to Object 1. %, which appears easier across multiple methods. 
 This trend is also observable in the results of instance-level methods, indicating that object-specific factors contribute in pose estimation accuracy. The domain shift between training and test data is a persistent issue with objects appearing in train scenes with more complex background and clutter to exhibit better accuracy at test time %(see Object 1 and 2 in Table~\ref{tab:ar_per_object}).
The relatively low performance of DOPE can be attributed to its reliance solely on image data and the corresponding projected bounding cuboids, without utilizing CAD models. %This approach necessitates a large and diverse training dataset that captures the object from numerous viewpoints to ensure robust performance. 
While CAD models could be used to generate additional synthetic data and increase viewpoint coverage and scene diversity, as proposed by the authors, we chose not to incorporate such data to maintain a consistent evaluation across all compared methods.
%Results with additional synthetic data are reported in the supplementary. For the case of novel object methods, the variants with the ground truth masks are reported in the Tables below as they exhibited better performance compared to the CNOS predictions.
%For a detailed breakdown of AR results per scene, showcasing the differences in scene configuration, their impact on performance and qualitative results, please refer to the supplementary material.

\paragraph{Detection and segmentation}
The detection and segmentation results are summarized in Table~\ref{tab:ap_metrics}. Both CNOS~\cite{nguyen2023cnos} and SAM-6D~\cite{Lin_2024_CVPR} achieve moderate performance, with consistently higher AP values on the extended dataset compared to the classic set. The main difference between the methods arises in segmentation on the classic set, where SAM-6D shows a clear advantage over CNOS, consistent with its SAM-based design for more precise mask generation. These findings highlight the complementary strengths of the two approaches and underline the continued challenges of accurate object localization in cluttered industrial scenes.

\section{Conclusions}
\label{sec:conclusions}
The IndustryShapes dataset features five challenging industrial components and tools and supports a wide range of evaluation scenarios, from single-object scenes to complex, cluttered environments with occlusions and multiple objects. The dataset is divided into two parts: the classic set and the extended set. The extended set is designed to support the evaluation of novel object pose estimation methods, particularly for industrially relevant objects, and includes RGB-D static onboarding sequences. Representative state-of-the-art methods were evaluated and the results demonstrate the strengths and current limitations of these methods in the industrial-related domain. \newline
\textbf{Limitations and future work.}~Some test scenes in the classic set comprise of fewer frames than others which may lead to uneven object representation across the dataset. The discrepancy in training scene complexity, with some objects trained on cluttered, realistic setups and others on simpler ones affects the performance of instance-level methods. These limitations suggest future improvements in increasing the number of frames in underrepresented test scenes, diversifying training data across all objects, and ensuring fuller pose coverage. %Moreover, the observed performance in novel-object pose estimation underscores the importance of a comprehensive review of existing methods in this domain, which could help guide future advancements in this challenging area.

%%%%%%%%%%%%%%%%%%%%%%%%%%%%%%%%%%%%%%%%%%%%%%%%%%%%%%%%%%%%%%%%%%%%%%%%%%%%%%%%

%References are important to the reader; therefore, each citation must be complete and correct. If at all possible, references should be commonly available publications.

\begin{table}[ht]
\caption{Mean Average Precision (mAP) of baseline methods on IndustryShapes dataset for detection and segmentation.}
\label{tab:ap_metrics}
\centering
\begin{tabular}{lcc|cc}
\toprule
 & \multicolumn{2}{c|}{\textbf{BBox – mAP}} & \multicolumn{2}{c}{\textbf{Segm – mAP}} \\
\cmidrule(lr){2-3} \cmidrule(lr){4-5}
\textbf{Method} & \textbf{Classic} & \textbf{Extended} & \textbf{Classic} & \textbf{Extended} \\
\midrule
CNOS   & 0.240 & \textbf{0.574} & 0.203 & \textbf{0.512} \\
SAM6D  & \textbf{0.270} & 0.527 & \textbf{0.345} & 0.453 \\
\bottomrule
\end{tabular}
\end{table}

%\section*{Acknowledgments}
%This work was funded by the European Union’s Horizon 2020 programme FELICE (GA No 101017151) European Union’s Horizon Europe programme SOPRANO (GA No 101120990). Special thanks to Stellantis—Centro Ricerche FIAT (CRF)/ SPW Research \& Innovation department in Melfi, Italy for supporting the dataset collection.

\bibliographystyle{IEEEtran}
\bibliography{references}

\end{document}